\documentclass[runningheads]{llncs}
\usepackage{esvect}
\usepackage[T1]{fontenc}
\usepackage{graphicx,verbatim}
\usepackage{amsmath}
\usepackage{amssymb}
\usepackage{multirow}

\usepackage{booktabs}
\usepackage{multirow}
\usepackage{graphicx}
\usepackage{array}
\usepackage{adjustbox}

\usepackage{dsfont}

\usepackage{float}

\newcommand{\grp}[1]{\rotatebox[origin=c]{90}{\scriptsize\textbf{#1}}}
\newcommand{\dshead}[1]{\textbf{#1}\;{\normalfont\small(Syn $\rightarrow$ Corrupted syn)}}

\usepackage{svg}

\usepackage{hyperref}
\usepackage{color}

\begin{document}
\title{Test Time Adaptation Methods for Point Cloud Registration in Laparoscopic Surgery}
\titlerunning{TTA for Point Cloud Registration in Laparoscopic Surgery}
%

\author{Nina Bodelot \and Soufiane Belharbi \and Eric Granger}  
\authorrunning{N. Bodelot et al.}
\institute{LIVIA, Dept. of Systems Engineering, ETS Montreal, Canada \\
    \email{nina.bodelot.1@ens.etsmtl.ca}, \email{eric.granger@etsmtl.ca}}
  
\maketitle              
\begin{abstract}
3D point cloud registration in laparoscopic surgery estimates the transformation between an intraoperative organ reconstructed from video and its preoperative mesh. Since ground-truth transformations are unavailable for real data, supervised networks are trained on synthetic organ pairs. At test time, however, real reconstructions differ from synthetic ones and are noisy, sparse and occluded, degrading correspondence estimation. Test-time adaptation (TTA) methods can address this domain shift at inference and have been applied successfully on tasks such as classification and segmentation. However, these methods typically rely on logits, entropy, class prototypes, or cache memory mechanisms unavailable in registration. Moreover, TTA methods assume a single shifted input, whereas registration involves a pair with an asymmetric shift, which mainly affects the intraoperative cloud, making TTA for registration much more challenging. This paper provides state-of-the-art TTA methods for 3D registration tasks across three families: model, normalization, and input adaptation. In particular, we analyze and modify four representative methods from those families that perform: auxiliary-task model update, backpropagation-free token purging, feature alignment, and layer-normalization calibration. TTA methods are modified to account for the asymmetric domain shifts between the preoperative and intraoperative point clouds, and to replace the entropy-based mechanism from classification. Given a correspondence-based model trained on clean source synthetic data, two scenarios were considered for the target: synthetic data with corruption and real data. We experiment on the P2P and P2ILReg datasets. Eight corruptions were applied separately to those datasets on the synthetic target data only, such as uniform noise and global density decrease, with five increasing levels of severity. Our experiments show that all methods improve registration on the P2P dataset, whereas on P2ILReg only input adaptation reduces the error, while normalization adaptation degrades it. Given the computational overhead of the backpropagation-based method, input adaptation is a more promising family for laparoscopic surgery, with low inference latency and a consistent reduction in error across all datasets.\\
Our code: \url{https://github.com/ninaa-git/survey_pc_registration_tta}
\end{abstract}

\section{Introduction}

Laparoscopic registration enhances surgical visualization by aligning a preoperative 3D organ mesh with a partial intraoperative surface reconstructed from 2D video frames \cite{huang_landmark-free_2025,ma_visualization_2023,zhou_landmark-free_2025}. Supervised models establish correspondences to estimate rigid transformations \cite{huang_landmark-free_2025,yao_pare-net_2025,zhou_landmark-free_2025} but require ground-truth labels unavailable during surgery. Source models are therefore trained on synthetic datasets and evaluated on real target data \cite{pan_survey_2010}. However, synthetic and real point clouds can differ in point density, visible surface, organ scale, and noise, introducing a domain shift between source and target. This shift can lead to incorrect registration, misleading the surgeon. Domain adaptation methods aim to mitigate such simulated-to-real distribution shift.
Source-free domain adaptation (SFDA) uses unlabeled target data without source access \cite{fang_source-free_2024,guichemerre_source-free_2024,sharafi_personalized_2026}. Despite its success, it remains impractical for laparoscopy as surgical data is unavailable for adaptation. In contrast, TTA adapts the model during inference from the incoming target data \cite{liang_comprehensive_2025,xiao_beyond_2024}, making it more suitable for surgery. We consider three families: model adaptation updates model parameters by backpropagation \cite{gidaris_unsupervised_2018}, normalization adaptation adapts normalization statistics \cite{wang_tent_2021}, and input adaptation modifies test embeddings before prediction \cite{dastmalchi_test-time_2024}. We study four representative methods of these families: Point-TTA \cite{hatem_point-tta_2023} for model adaptation,
Layer Normalization (LN) \cite{yazdanpanah_revisiting_nodate} for normalization adaptation, Progressive Embedding Alignment (PEA) \cite{ma_architecture-agnostic_2026} and Purge-Gate \cite{yazdanpanah_purge-gate_nodate} for input adaptation.
Despite its suitability, TTA remains limited for registration. Most methods were designed for classification \cite{wang_backpropagation-free_2024,wei_3d_nodate} or segmentation \cite{saltori2022gipso}, relying on logits, entropy, class prototypes, or cache memory, limiting their transfer for rigid transformation tasks. Applying these TTA methods to registration therefore requires changes. 
First, classification TTA adapts one target input. In registration, the target is a pair in which the shift mainly affects the intraoperative cloud. A single-input method estimates one correction and applies it to the whole pair, introducing a shift in the preoperative cloud. We therefore modify each method to account for this asymmetry.
Then, logit-based criterion from classification do not transfer to registration tasks that output a transformation. It is therefore replaced.


\noindent \textbf{Key contributions:}
    \textbf{(1)} 
    An analysis and modification of four representative methods across three TTA families (Point-TTA, LN, PEA, Purge-Gate) for 3D registration. They are modified to account for the asymmetric domain shift between the preoperative and intraoperative point clouds. We also replace the classification entropy-based mechanism of Purge-Gate with an inlier ratio mechanism.
    \textbf{(2)} 
    A comparative evaluation of four TTA methods for laparoscopic point cloud registration on P2P and P2ILReg datasets \cite{yang_resolving_2025,zhou_landmark-free_2025}, with Point-TTA on P2P only. Experiments are performed under two target-domain scenarios (corrupted synthetic and real intraoperative data), under episodic and continual settings. Our results show that the adaptation methods consistently minimize registration error on the P2P dataset, whereas input adaptation reduces the error, and normalization adaptation degrades performance on the P2ILReg dataset. Model adaptation introduces high computational overhead. Input adaptation family improves performance across all datasets and has a low inference latency, making it more suitable for laparoscopy surgery. 

\section{Related Work}

\noindent \textbf{(a) Point Cloud Registration:}
Deep rigid registration models extract and match embeddings to estimate global transformations \cite{zhang_deep_2026}. Among these models, the supervised frameworks are broadly split into one-stage and two-stage architectures. One-stage methods directly estimate the transformation from the input points, whereas two-stage methods first match superpoints before computing the global transformation \cite{yang_3d_2024}. 
For laparoscopy surgery, both one-stage registration networks and superpoint-based methods have been proposed \cite{yang_learning_2023,zhou_landmark-free_2025}. Among supervised methods, PARE-Net \cite{yao_pare-net_2025} is an effective approach \cite{zhang_deep_2026}, which is two-stage. It extracts point embeddings with specific convolutions and uses transformers to improve point-cloud matching \cite{yao_pare-net_2025}.

\noindent \textbf{(b) Test-Time Adaptation for 3D Point Cloud Registration:}
Test-time adaptation methods have mainly been explored in 2D \cite{wang_tent_2021}, and they underperform when directly applied to 3D settings \cite{leonardis_cloudfixer_2025}. While 3D methods have mostly focused on classification \cite{wang_backpropagation-free_2024,wei_3d_nodate}, only one method specifically addresses point cloud registration \cite{hatem_point-tta_2023}. Existing TTA approaches can be classified into four families according to the adapted component: the weights, normalization layers, input or predictions \cite{liang_comprehensive_2025,xiao_beyond_2024}. They can also be episodic \cite{hatem_point-tta_2023,liang_comprehensive_2025}, where each batch is adapted independently, or online \cite{ma_architecture-agnostic_2026,xiao_beyond_2024}, where the adaptation from previous target batches is reused for the following batches. 

The first family, model adaptation, relies on auxiliary tasks or on test-time objectives. 
Auxiliary-task methods update an encoder shared between the main task and a self-supervised task, such as rotation prediction \cite{gidaris_unsupervised_2018}, contrastive learning \cite{alet_tailoring_2021}, or reconstruction \cite{hatem_point-tta_2023,jiang_pointmac_2025,pang_masked_2022}. 
Objective-based methods directly update the model's weights at test time by minimizing prediction entropy \cite{wang_tent_2021}, enforcing feature alignment \cite{kojima_robustifying_2022}, or optimizing against pseudo-labels \cite{sun_point-cache_nodate}. Then, the normalization family updates only layer or batch normalization statistics to realign the target data with the source domain knowledge embedded in these layers \cite{wang_tent_2021,bahri_smart-pc_2025,bahri_test-time_2025,yazdanpanah_revisiting_nodate}. Although lightweight, this strategy requires accurate target distribution estimation that has been best achieved in continual settings with strategies such as weighting \cite{hu_mixnorm_2021} or class-balanced cache \cite{3600270.3602246}.
%
The input adaptation family aligns target inputs or embeddings with the source domain \cite{dastmalchi_test-time_2024,leonardis_cloudfixer_2025}. This improves source model performance as the input or embeddings resemble the source data distribution. 
While it avoids any model updates and can be applied to all model architectures, it can fail under severe domain shifts if the input adaptation is insufficient.
Inference adaptation modifies the model output, for example through caching using logits \cite{sun_point-cache_nodate} or combining different model predictions \cite{wang_efficient_2021,zhang_domain-specific_2023}. While this method improves the output predictions, it leaves the input and model untouched, which fails under drastic data shifts.

However, for all families, most of these methods are designed for classification and segmentation tasks \cite{liang_comprehensive_2025,sun_point-cache_nodate}. The direct application of those families to rigid registration remains limited, having been attempted in model adaptation with 3D auxiliary tasks \cite{hatem_point-tta_2023} with indoor and outdoor scenes.

\begin{figure}[!t]
    \centering
    \includegraphics[width=.85\linewidth]{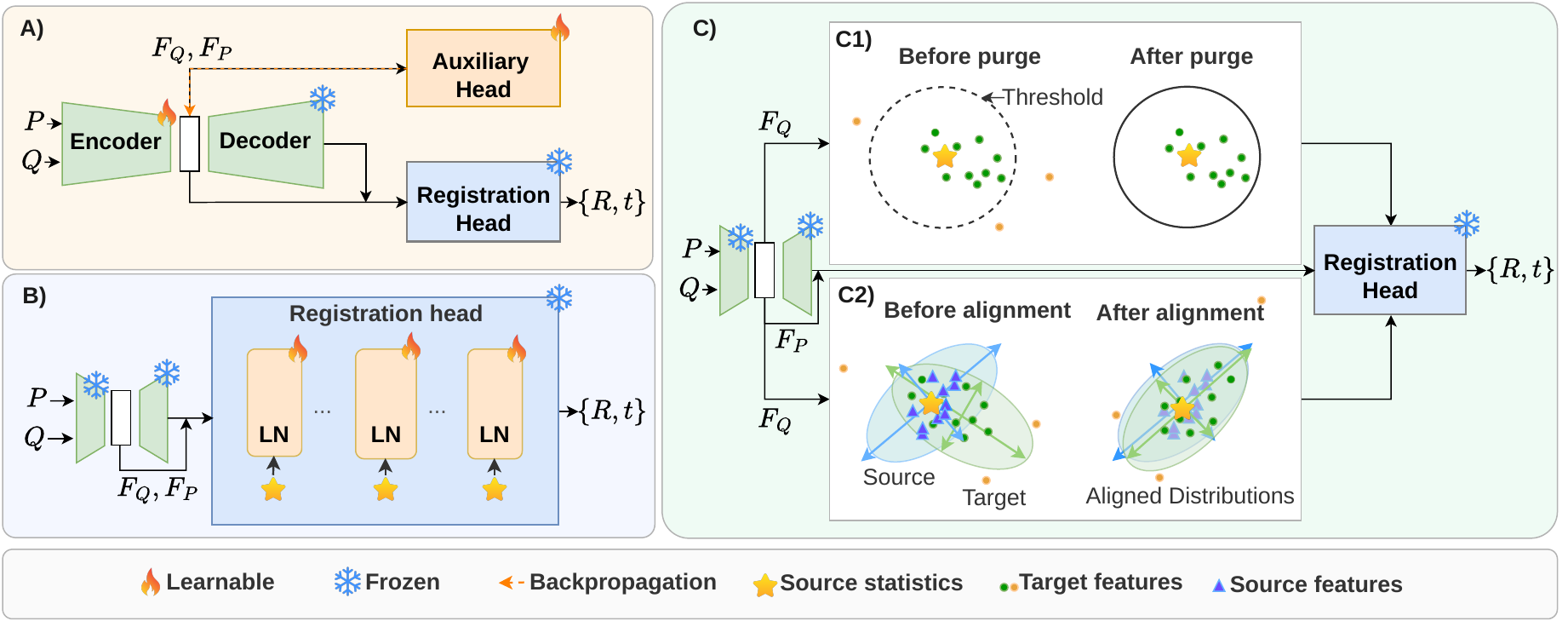}
    \caption{
    Modified TTA methods, with $P, Q$: preoperative and intraoperative point clouds, $\mathcal{T}=\{R,t\}$: transformation, $F_P,F_Q$: output of the encoder.   
    (A) Model adaptation using Point-TTA \cite{hatem_point-tta_2023} relies on an auxiliary head. 
    (B) Normalization adaptation modifies the normalization mechanism, for LN \cite{yazdanpanah_revisiting_nodate} using source and target statistics. 
    (C) Input adaptation before the registration head: (C1) Purge-Gate \cite{yazdanpanah_purge-gate_nodate} removes the divergent target features, (C2) PEA \cite{ma_architecture-agnostic_2026} aligns target features to source statistics.
    }
    \label{fig:placeholder}
\end{figure}

\section{TTA Methods for Registration}

This section describes our modifications to representative methods for registration (Fig.~\ref{fig:placeholder}). 
In laparoscopic surgery, the shift mainly affects the intraoperative point cloud, whereas the preoperative point cloud is processed like the source mesh and is far less affected.

\noindent \textbf{A) Point-TTA \cite{hatem_point-tta_2023}} is a model adaptation method allowing the update of weights at test time. The registration model contains a shared feature encoder, a primary registration branch, and an auxiliary branch. The shared encoder is adapted with backpropagation using self-supervised auxiliary losses. These losses are defined from auxiliary tasks such as point cloud reconstruction, feature learning, and correspondence classification \cite{hatem_point-tta_2023}. 
Point-TTA uses a joint main and auxiliary task training for initialization, followed by a meta-auxiliary learning strategy \cite{finn_model-agnostic_2017} so that the auxiliary update benefits the primary registration task. At test time, the auxiliary loss updates the shared backbone for each sample, while the registration head remains frozen since no target registration labels are available.
This enables sample-specific adaptation of the feature extractor. However, the source training is a joint stage followed by a meta training stage before deployment. Its main limitation for time-constrained registration is its high computational cost, needed for several forward and backward passes.

\textbf{Modifications:} The reconstruction auxiliary task is used since it is the most effective auxiliary task in registration \cite{hatem_point-tta_2023}. We perform a two-stage point cloud registration and use the encoder embeddings of the coarse point matching for coarse point cloud reconstruction. 
Since the encoder is shared between the intraoperative and preoperative clouds, adaptation uses the embeddings of both intraoperative and preoperative point clouds rather than only those of the intraoperative cloud.

\noindent \textbf{B) Layer Normalization \cite{yazdanpanah_revisiting_nodate}:}
Normalization calibration methods restrict adaptation parameters of normalization layers, as they contain source scaling and shifting statistics.
In LN \cite{yazdanpanah_revisiting_nodate}, the pre-affine normalization is preserved, while the post-affine output is adapted using source and target moments. The target moments can be estimated using one batch only in an episodic setting, which can be unreliable, or over a window of batches.
Although rapid, this method limits the adaptation capacity.

\textbf{Modifications:} LN~\cite{yazdanpanah_revisiting_nodate} was designed for classification, where each LN layer processes a single input and stores one source statistic. In registration, both the preoperative and intraoperative clouds pass through the same LN layers, in self-attention within each cloud and in cross-attention between them, but the shift mainly affects the intraoperative one. A statistic shared across the two clouds would therefore re-shift the preoperative cloud toward the intraoperative one. We instead split each adapted LN layer's statistics by cloud, storing separate source and target moments for each, and calibrate every cloud 
with its own. The target moments are estimated online. 

\noindent \textbf{C) Input Adaptation:}
This family adapts the input data before processing it while leaving the model unchanged ((C) in Fig.\ref{fig:placeholder}). For instance, Purge-Gate \cite{yazdanpanah_purge-gate_nodate} (C1 in Fig.\ref{fig:placeholder}) removes tokens whose embeddings are the most divergent from a source-domain prototype. Removing outlier tokens allows the model predictions to be more accurate.
The method is lightweight and episodic. However, the purging ratio is selected by minimizing classification entropy. This criterion is linked to classification logits and must therefore be modified for registration. Another work within this family is PEA \cite{ma_architecture-agnostic_2026}(C2 in Fig.\ref{fig:placeholder}). Instead of removing tokens, it aligns target embeddings toward source feature statistics. Source mean and covariance are computed offline, while target statistics are obtained online and used for covariance alignment.
The aligned embeddings are then passed to the frozen registration head. PEA is designed to update its target statistics in a continual mode that relies on an exponential moving average (EMA) with momentum.
PEA is backpropagation-free and does not modify the model weights. However, the computation of covariance matrices and their decomposition can introduce additional cost, especially when applied at several layers.

\textbf{Modifications:} Adaptation is applied to the embeddings of the intraoperative point cloud only, as it is the point cloud that contains the main shift. For Purge-Gate\cite{yazdanpanah_purge-gate_nodate}, we replace the entropy-based selection with the Inlier Ratio (IR), a common unsupervised measure of correspondence quality in registration \cite{yang_3d_2024,zhang_deep_2026}. We select the purging ratio $\mathcal{L}_{pg}$ that maximizes
$\mathrm{IR}(\hat{T},\hat{\mathcal{C}})=\frac{1}{k}\sum_{i=1}^{k}\mathds{1}(\|\hat{R}\mathbf{p}_i+\hat{t}-\mathbf{q}_i\|_2<\tau)$,
where $(\mathbf{p}_i,\mathbf{q}_i)$ are matched correspondences and $\tau$ is the acceptance radius. 
For PEA, as for Purge-Gate, we only adapt the embeddings associated with the intraoperative point cloud. This differs from the original PEA, where the encoder is adapted progressively. In our case, adapting the encoder is not possible as the alignment would also shift the embeddings of the preoperative point cloud, which is far less affected by the shift.

For PEA \cite{ma_architecture-agnostic_2026} and LN \cite{yazdanpanah_revisiting_nodate}, we consider statistics estimated either with an episodic (EP) strategy, where statistics are estimated per sample or a continual strategy (CT) where statistics are accumulated across samples. Both methods are designed for continual adaptation 
as statistical estimation with one sample is unreliable. LN averages statistics over a window \cite{yazdanpanah_revisiting_nodate}, and PEA uses EMA of step $m$ \cite{ma_architecture-agnostic_2026}. For a fair comparison, we use a modified EMA for both LN and PEA. In the EMA formulation, the first batch estimation is biased when it contains one sample. Let $n$ be the first cumulative number of embeddings. We compute the target statistics as a cumulated mean \cite{welford1962}, and switch to EMA once its update falls below the momentum $m$.

\section{Results and Discussion}\label{results-discussion}

\subsection{Experimental methodology}

\noindent \textbf{Datasets and corruptions:}
Two synthetic datasets are used for source model training, and three target sets for testing (two corrupted synthetic, one real). The first synthetic dataset, P2P \cite{yang_resolving_2025}, contains preoperative livers 
and intraoperative liver that are cropped to visible surfaces of the preoperative liver. 
We split the original training set into 80\%/20\% for training and validation for 1048/261 samples, and 
test on the provided 6315 liver pairs. The second one, P2ILReg \cite{zhou_landmark-free_2025}, provides synthetic and real test sets
with preoperative meshes. The real test set has 92 intraoperative point clouds reconstructed from individual laparoscopic video key frames.
The synthetic set is a 60\%/20\%/20\% train/validation/test split, provided by \cite{zhou_landmark-free_2025} and totaling 29400/9800/9800 samples. 
To emulate target-domain shifts, corruptions are applied only to the synthetic 
intraoperative target point clouds. The real test set remains unchanged, 
since real data inherently introduces a domain shift. We use noise corruptions: uniform (uni), Gaussian (gauss), background (backg), impulse (impul); and sampling/visibility corruptions: global density decrease (g-d-dec), local density decrease (l-d-dec), cutout (cut), occlusion (oclsion). Each corruption is generated at five levels of increasing severity, following 
point cloud corruption benchmarks \cite{sun_benchmarking_2022} and laparoscopic liver registration settings \cite{zhou_landmark-free_2025}, with results averaged over severity levels. \\
\noindent \textbf{Evaluation metrics:}
For synthetic data, we report relative rotation error (RRE) and relative translation error (RTE), following \cite{zhou_landmark-free_2025}. We additionally report root mean squared error (RMSE) directly, rather than the registration recall used in \cite{zhou_landmark-free_2025} to avoid thresholding RMSE at a specific distance.
For real intraoperative data, ground-truth rigid transformations are not available. Following \cite{zhou_landmark-free_2025}, we report Chamfer Distance (CD)\cite{barrow_parametric_1977} between 
the point clouds, and the Dice score between the projected registered organ surface and the annotated organ mask.\\
\noindent \textbf{Implementation details:}
PARE-Net \cite{yao_pare-net_2025} is used as the correspondence-based registration backbone. 
Following the dataset protocols \cite{yang_resolving_2025,zhou_landmark-free_2025}, training is performed for 150/80 epochs on P2P/P2ILReg.
Batch size is 12 for source training and 1 for testing, as key frames are processed one at a time.
Experiments are run on NVIDIA H100 (P2P) and NVIDIA RTX A6000 (P2ILReg). 
For Point-TTA, the learning rate for the auxiliary task is  $\gamma=1.6\times10^{-4}$, and $K=5$ steps to update the backbone as in \cite{hatem_point-tta_2023}. Point-TTA is only evaluated on the synthetic P2P data \cite{yang_resolving_2025} as its joint and meta training on P2ILReg is computationally prohibitive. For modified methods, TTA hyperparameters are selected per dataset by minimizing RMSE on the source validation set rather than the corrupted test data. This avoids test label leakage, but may be suboptimal since the validation data is clean. 
For Purge-Gate \cite{yazdanpanah_purge-gate_nodate}, the purging ratio is selected among $\mathcal{L}_{pg}\in\{0.0,0.1,0.5\}$. For continual mode, PEA uses $m=0.05$ for P2P dataset \cite{yang_resolving_2025} and $m=0.02$ for P2ILReg \cite{zhou_landmark-free_2025}, and LN uses $m=0.01$ on both datasets.

\subsection{Comparison of State-of-the-Art Methods}

\begin{figure}[!t]
    \centering
    \includegraphics[width=0.85\linewidth]{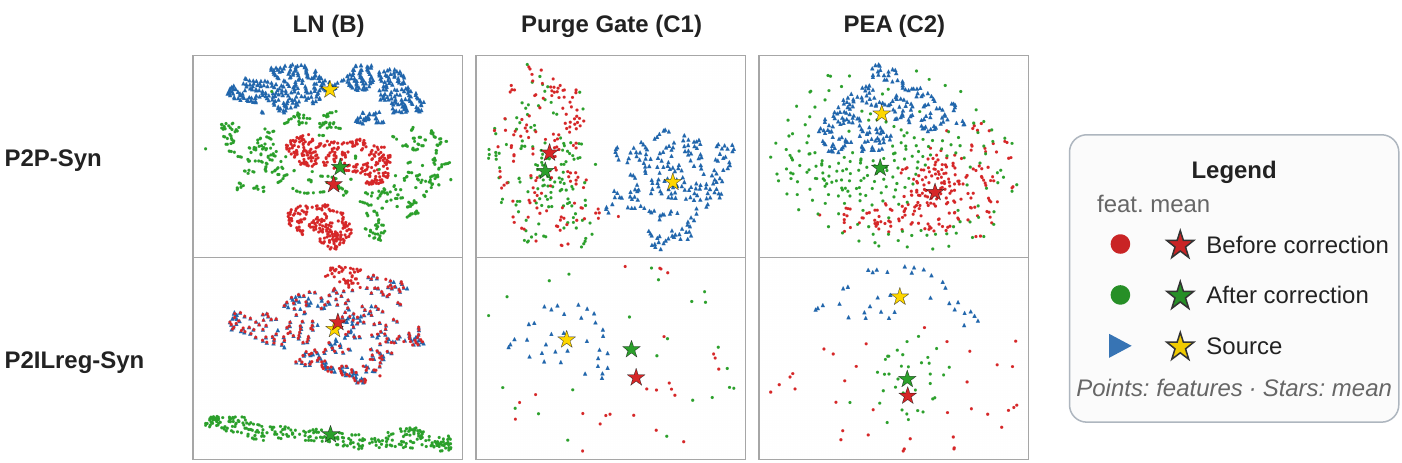} 
    \caption{t-SNE for uniform noise 5. For LN and PEA, embeddings appear before (red) and after (green) correction. For Purge-Gate, tokens are purged (red) or kept (green).} 
    \label{fig:tsne}
\end{figure}
Table~\ref{tab:results_rmse} shows that all modified methods improve registration on P2P-Syn, led by LN ($-$5.41 episodic, $-$5.50 continual). On the more challenging P2ILReg-Syn dataset, whose source baseline is higher (38.70 mm vs 15.47 mm), LN degrades registration ($+$8.61). Indeed, under large shifts, LN target statistics become unreliable and move features away from the source. PEA and Purge-Gate remain robust ($-$1.80, $-$1.63). PEA registration is degraded notably for background noise on both datasets. Purge-Gate improves registration on both datasets and for any corruption, as it selects the purge ratio by inlier ratio and keeps the unpurged input when purging would lower it. 
Table \ref{tab:results_rre_rte} confirms with RRE/RTE that PEA and Purge-Gate improve registration across the synthetic datasets. However, the gains remain small, around 2° and 1mm, which is limited for surgery relative to the source-only errors of 21°/15mm on P2P-Syn and 36°/24mm on P2ILReg-Syn. 
On P2ILReg-Real, LN again degrades registration. Purge-Gate and continual PEA improve registration on CD/Dice, from 4.26mm/71.87 to 4.17mm/72.97 and 4.22mm/72.22, respectively. PEA episodic is close to the source and improves Dice to 72.85. As with the synthetic data, improvements remain limited.
Unlike LN, Purge-Gate and continual PEA improve registration over all datasets. On synthetic datasets, the continual mode improves LN while improving (P2ILReg-Syn) or degrading (P2P-Syn) PEA. On the real dataset, it degrades LN; and on PEA improves CD by 0.07 mm and degrades the Dice score by 0.67. Samples on synthetic and real datasets come from unordered samples or non-consecutive frames, respectively, from different patients. Accumulated target statistics may therefore be too disparate to yield accurate estimates.

\noindent\textbf{t-SNE visualizations:} 
Fig.~\ref{fig:tsne} suggests a limited adaptation, yielding a marginal shift toward the source mean. Purge-Gate only prunes points and cannot align them (Fig.~\ref{fig:tsne}). For PEA and LN, suboptimal alignment could come from inaccurate target statistic estimation. To address this, input adaptation methods could be extended with progressive alignment or purge across the encoder and decoder. Moreover, since PEA or LN accumulated statistics can become outdated, continual estimation methods~\cite{gao_apcotta_2026,jiang_pcotta_nodate} could improve them.

\begin{table}[!t]
\centering
\scriptsize
\setlength{\tabcolsep}{2pt}
\resizebox{.8\linewidth}{!}{%
\begin{tabular}{cl|cccccccc|c}
\toprule
& \textbf{Method}
& uni & gauss & backg & impul
& g-d-dec & l-d-dec & cut & oclsion & \textbf{Mean} \\
\midrule
\multicolumn{11}{c}{\dshead{P2P-Syn}} \\
\midrule
& Source only & 28.28 & 32.24 & 6.99 & 15.91  & 17.21 & 10.68 & 5.95 & 6.48 & 15.47 ± 30.14 \\
\cmidrule(lr){2-11}
\multirow{4}{*}{\grp{EP}}
& Point-TTA~\cite{hatem_point-tta_2023}\,{\tiny(ICCV'23)} & 24.96 & 27.57 & 6.46 & 14.91 & 23.15 & 9.43 & 5.30 & 6.18 &  14.75 ± 29.55 (-0.72)\\
& Purge-Gate~\cite{yazdanpanah_purge-gate_nodate}\,{\tiny(ICCV'25)} & 27.71 & 31.78 & 4.41 & 14.19 & 16.23 & 7.29 & 4.30 & 4.46 & 13.79 ± 28.35 (-1.68)\\
& LN~\cite{yazdanpanah_revisiting_nodate}\,{\tiny(WACV'26)} & 15.18 & 16.83 & 6.89 & 10.13 & 9.87 & 9.12 & 5.93 & 6.54 & 10.06 ± 22.43 (-5.41) \\
& PEA~\cite{ma_architecture-agnostic_2026}\,{\tiny(ICLR'26)} & 21.80 & 24.17 & 9.20 & 14.01 & 12.40 & 11.84 & 8.40 & 8.84 & 13.83 ± 27.66 (-1.64)\\
\cmidrule(lr){2-11}
\multirow{2}{*}{\grp{CT}}
& LN~\cite{yazdanpanah_revisiting_nodate}\,{\tiny(WACV'26)} & 15.29 & 16.95 & 6.78 & 9.80 & 10.09 & 8.81 & 5.72 & 6.34 &  9.97 ± 22.38(-5.50) \\
& PEA~\cite{ma_architecture-agnostic_2026}\,{\tiny(ICLR'26)} & 22.93 & 25.46 & 7.93 & 13.67 & 15.46 & 10.98 & 7.39 & 8.05 & 13.98 ± 27.32(-1.49)\\
\midrule
\addlinespace[2pt]
\multicolumn{11}{c}{\dshead{P2ILReg-Syn}} \\
\midrule
& Source only & 88.31 & 104.24 & 23.49 & 82.09 & 3.35 & 2.23 & 2.85 & 3.07 & 38.70 ± 57.98 \\
\cmidrule(lr){2-11}
\multirow{4}{*}{\grp{EP}}
& Purge-Gate~\cite{yazdanpanah_purge-gate_nodate}\,{\tiny(ICCV'25)} & 87.37 & 103.11 & 17.54 & 80.85 & 2.36 & 1.52 & 1.85 & 1.98 & 37.07 ± 57.57
 (-1.63)\\
& LN~\cite{yazdanpanah_revisiting_nodate}\,{\tiny(WACV'26)} & 97.73 & 115.38 & 41.69 & 97.08  & 7.83 & 5.67 & 6.50 & 6.60 & 47.31 ± 62.61 (+8.61)\\
& PEA~\cite{ma_architecture-agnostic_2026}\,{\tiny(ICLR'26)} & 84.84 & 100.28 & 25.35 & 73.23 & 3.35 & 2.25 & 2.81 & 3.05 & 36.90 ± 56.48 (-1.80)\\
\cmidrule(lr){2-11}
\multirow{2}{*}{\grp{CT}}
& LN~\cite{yazdanpanah_revisiting_nodate}\,{\tiny(WACV'26)} & 97.03 & 114.25 & 41.20 & 95.74 & 7.00 & 5.02 & 5.72 & 5.90 & 46.48 ± 61.80 (+7.78)\\
& PEA~\cite{ma_architecture-agnostic_2026}\,{\tiny(ICLR'26)} & 84.49 & 99.59 & 25.50 & 73.14 & 3.35 & 2.19 & 2.81 & 3.04 & 36.77 ± 56.20 (-1.93)\\
\bottomrule
\end{tabular}%
}
\caption{Mean RMSE ± standard deviation (mm) of modified TTA methods on P2P-Syn and P2ILReg-Syn datasets. Continual (CT) and episodic (EP) denote the statistical estimation strategies. }
\label{tab:results_rmse}
\end{table}

\begin{table}[!t]
\centering
\scriptsize
\setlength{\tabcolsep}{2pt}
\resizebox{.9\linewidth}{!}{%
\begin{tabular}{cl|cc|cc|cc}
\toprule
& & \multicolumn{2}{c|}{\textbf{P2P-Syn}} & \multicolumn{2}{c}{\textbf{P2ILReg-Syn}} & \multicolumn{2}{c}{\textbf{P2ILReg-Real}}\\
& & \multicolumn{2}{c|}{Syn $\rightarrow$ Corrupted syn} & \multicolumn{2}{c}{Syn $\rightarrow$ Corrupted syn} & \multicolumn{2}{c}{Syn $\rightarrow$ Real}\\
\cmidrule(lr){3-4} \cmidrule(lr){5-6} \cmidrule(lr){7-8}
& \textbf{Method}
& RRE ($^\circ$)\,$\downarrow$ & RTE (mm)\,$\downarrow$
& RRE ($^\circ$)\,$\downarrow$ & RTE (mm)\,$\downarrow$ 
& CD (mm)\,$\downarrow$ & Dice\,$\uparrow$ \\
\midrule
& Source only & 21.34 ± 46.87 & 14.81 ± 75.53 & 36.13 ± 56.33 &  24.25 ± 39.68 & 4.26 ± 1.82 & 71.87 ± 12.61 \\
\cmidrule(lr){2-8}
\multirow{4}{*}{\grp{EP}}
& Point-TTA~\cite{hatem_point-tta_2023}\,{\tiny(ICCV'23)} & 19.58 ± 44.68 & 13.79 ± 69.18 & -- & -- & -- & -- \\
& Purge-Gate~\cite{yazdanpanah_purge-gate_nodate}\,{\tiny(ICCV'25)} & 18.89 ± 44.26 & 12.50 ± 65.60 & 34.69 ± 56.38 & 22.90 ± 38.85 & 4.17 ± 1.57 & 72.97 ± 10.85 \\
& LN~\cite{yazdanpanah_revisiting_nodate}\,{\tiny(WACV'26)} & 12.83 ± 34.61 & 11.43 ± 64.21 & 43.09 ± 59.60 & 30.11 ± 44.75 & 5.34 ± 2.41  & 67.50 ± 14.51 \\
& PEA~\cite{ma_architecture-agnostic_2026}\,{\tiny(ICLR'26)} & 18.08 ± 42.47 & 13.65 ± 68.48
 & 34.19 ± 54.78 & 22.90 ± 39.00 & 4.27 ± 1.78 & 72.85 ± 12.18 \\
\cmidrule(lr){2-8}
\multirow{2}{*}{\grp{CT}}
& LN~\cite{yazdanpanah_revisiting_nodate}\,{\tiny(WACV'26)} & 12.80 ± 34.60 & 11.30 ± 64.07 & 42.61 ± 59.27 & 28.94 ± 43.57 & 5.59 ± 2.80 & 65.52 ± 15.17 \\
& PEA~\cite{ma_architecture-agnostic_2026}\,{\tiny(ICLR'26)} & 18.12 ± 41.70 & 14.29 ± 69.48 & 34.20 ± 54.73 & 22.48 ± 38.29 & 4.20 ± 1.45  & 72.18 ± 11.42 \\
\bottomrule
\end{tabular}
}
\caption{Mean RRE($^\circ$) and RTE(mm) ± standard deviation on synthetic datasets, and CD(mm) and Dice ± standard deviation on the real dataset.}
\label{tab:results_rre_rte}
\end{table}

\noindent \textbf{Computational cost:} 
On the P2P dataset, the source-only average time per sample is 162 ms, and it is 184/189/414 ms for LN/PEA/Purge-Gate, compared to 1226 ms for Point-TTA. Peak GPU memory for source-only and the backpropagation-free methods stays at 1128–1187 MB, but reaches 5370 MB for Point-TTA. Since registration adaptation must be lightweight, this efficiency is decisive, and Point-TTA is less suited for real-time surgery.

\section{Conclusion}
Domain shift between synthetic training data and intraoperative reconstructions degrades laparoscopic registration. We analyzed and modified for registration four TTA methods across model, normalization, and input adaptation families. On corrupted synthetic and real data, input adaptation reduces registration errors, remaining the most robust family, with smaller and mixed gains on real data. Conversely, normalization adaptation only helps P2P, while model adaptation introduces computational overhead. This suggests that the TTA input adaptation family improves registration and could be further explored.

\noindent\textbf{Acknowledgments.} This work was supported in part by INOVAIT, the Government of Canada's Strategic Innovation Fund, and the Natural Sciences and Engineering Research Council of Canada. The Digital Research Alliance of Canada provided the computing resources. We thank Sébastien Delorme and his team at Scopia Surgical for contributing to the problem formulation.\par


%
%
%

\bibliographystyle{splncs04}
\bibliography{biblio}

@article{yang_resolving_2025,
	title = {Resolving the {Ambiguity} of {Complete}-to-{Partial} {Point} {Cloud} {Registration} for {Image}-{Guided} {Liver} {Surgery} with {Patches}-to-{Partial} {Matching}},
	journal = {IEEE JBHI},
	author = {Yang, Zixin and Heiselman, Jon S. and Han, Cheng and others},
	year = {2026},
}

@inproceedings{wu_density-aware_2021,
  title={Density-aware Chamfer Distance as a Comprehensive Metric for Point Cloud Completion},
  author={Wu, Tong and Pan, Liang and Zhang, Junzhe and Wang, Tai and Liu, Ziwei and Lin, Dahua},
  booktitle={NeurIPS},
  year={2021}
}

@article{zhou_landmark-free_2025,
	title = {Landmark-{Free} {Preoperative}-to-{Intraoperative} {Registration} in {Laparoscopic} {Liver} {Resection}},
        journal = {IEEE TMI},
	author = {Zhou, Jun and Gao, Bingchen and Wang, Kai and others},
	year = {2025},
}

@misc{yang_3d_2024,
	title = {{3D} {Registration} in 30 {Years}: {A} {Survey}},
	shorttitle = {{3D} {Registration} in 30 {Years}},
	author = {Yang, Jiaqi and Zhang, Chu'ai and Wang, Zhengbao and others},
	year = {2024},
}

@inproceedings{bahri_smart-pc_2025,
	title = {{SMART}-{PC}: {Skeletal} {Model} {Adaptation} for {Robust} {Test}-{Time} {Training} in {Point} {Clouds}},
	shorttitle = {{SMART}-{PC}},
	booktitle = {ICML},
	author = {Bahri, Ali and Yazdanpanah, Moslem and Dastani, Sahar and others},
	year = {2025},
}

@article{fang_source-free_2024,
	title = {Source-free unsupervised domain adaptation: {A} survey},
	journal = {Neural Networks},
	author = {Fang, Yuqi and Yap, Pew-Thian and Lin, Weili and Zhu, Hongtu and Liu, Mingxia},
	year = {2024},
}

@article{liang_comprehensive_2025,
	title = {A {Comprehensive} {Survey} on {Test}-{Time} {Adaptation} {Under} {Distribution} {Shifts}},
	journal = {IJCV},
	author = {Liang, Jian and He, Ran and Tan, Tieniu},
	year = {2025},
}

@inproceedings{yao_pare-net_2025,
	title = {{PARE}-{Net}: {Position}-{Aware} {Rotation}-{Equivariant} {Networks} for {Robust} {Point} {Cloud} {Registration}},
	booktitle = {ECCV},
	author = {Yao, Runzhao and Du, Shaoyi and Cui, Wenting and Tang, Canhui and Yang, Chengwu},
	year = {2024},
}

@inproceedings{hatem_point-tta_2023,
	title = {Point-{TTA}: {Test}-{Time} {Adaptation} for {Point} {Cloud} {Registration} {Using} {Multitask} {Meta}-{Auxiliary} {Learning}},
	booktitle = {ICCV},
	author = {Hatem, Ahmed and Qian, Yiming and Wang, Yang},
	year = {2023},
}

@misc{sun_benchmarking_2022,
	title = {Benchmarking {Robustness} of {3D} {Point} {Cloud} {Recognition} {Against} {Common} {Corruptions}},
	publisher = {arXiv},
	author = {Sun, Jiachen and Zhang, Qingzhao and Kailkhura, Bhavya and Yu, Zhiding and Xiao, Chaowei and Mao, Z. Morley},
	year = {2022},
}

@inproceedings{jiang_pointmac_2025,
	title = {{PointMAC}: {Meta}-{Learned} {Adaptation} for {Robust} {Test}-{Time} {Point} {Cloud} {Completion}},
	booktitle = {NeurIPS},
	author = {Jiang, Linlian and Ma, Rui and Gu, Li and Wang, Ziqiang and Zuo, Xinxin and Wang, Yang},
	year = {2025},
}

@misc{xiao_beyond_2024,
	title = {Beyond {Model} {Adaptation} at {Test} {Time}: {A} {Survey}},
	publisher = {arXiv},
	author = {Xiao, Zehao and Snoek, Cees G. M.},
	year = {2024},
}

@inproceedings{bahri_test-time_2025,
	title = {Test-{Time} {Adaptation} in {Point} {Clouds}: {Leveraging} {Sampling} {Variation} with {Weight} {Averaging}},
	booktitle = {WACV},
	author = {Bahri, Ali and Yazdanpanah, Moslem and Noori, Mehrdad and others},
	year = {2025},
}

@inproceedings{jiang_pcotta_nodate,
	title = {{PCoTTA}: {Continual} {Test}-{Time} {Adaptation} for {Multi}-{Task} {Point} {Cloud} {Understanding}},
        booktitle = {NeurIPS},
        year = {2024},
	author = {Jiang, Jincen and Zhou, Qianyu and Li, Yuhang and others},
}

@inproceedings{dastmalchi_test-time_2024,
	title = {Test-{Time} {Adaptation} of {3D} {Point} {Clouds} via {Denoising} {Diffusion} {Models}},
	booktitle = {WACV},
	author = {Dastmalchi, Hamidreza and An, Aijun and Cheraghian, Ali and others},
	year = {2025},
}

@inproceedings{wang_backpropagation-free_2024,
	title = {Backpropagation-free {Network} for {3D} {Test}-time {Adaptation}},
	booktitle = {CVPR},
	author = {Wang, Yanshuo and Cheraghian, Ali and Hayder, Zeeshan and others},
	year = {2024},
}

@article{huang_landmark-free_2025,
	title = {A {Landmark}-{Free} {3D}–{2D} {Rigid} {Liver} {Registration} via {Point} {Cloud} {Matching} for {Laparoscopic} {Surgery}},
	journal = {HTL},
	author = {Huang, Binyan and Yang, Xiangyue and Jia, Fucang},
	year = {2025},
}

@inproceedings{yazdanpanah_purge-gate_nodate,
	title = {Purge-{Gate}: {Backpropagation}-{Free} {Test}-{Time} {Adaptation} for {Point} {Clouds} {Classification} via {Token} purging},
	booktitle = {ICCV},
	author = {Yazdanpanah, Moslem and Bahri, Ali and Noori, Mehrdad and others},
	year = {2025},
}

@inproceedings{sun_point-cache_nodate,
	title = {Point-{Cache}: {Test}-time {Dynamic} and {Hierarchical} {Cache} for {Robust} and {Generalizable} {Point} {Cloud} {Analysis}},
	booktitle = {CVPR},
	author = {Sun, Hongyu and Ke, Qiuhong and Cheng, Ming and others},
	year = {2025},
}

@inproceedings{leonardis_cloudfixer_2025,
	title = {{CloudFixer}: {Test}-{Time} {Adaptation} for {3D} {Point} {Clouds} via {Diffusion}-{Guided} {Geometric} {Transformation}},
	booktitle = {ECCV},
	author = {Shim, Hajin and Kim, Changhun and Yang, Eunho},
	year = {2024},
}

@article{yang_learning_2023,
	title = {Learning {Feature} {Descriptors} for {Pre}- and {Intra}-operative {Point} {Cloud} {Matching} for {Laparoscopic} {Liver} {Registration}},
	journal = {IJCARS},
	author = {Yang, Zixin and Simon, Richard and Linte, Cristian A.},
	year = {2023},
}

@inproceedings{pang_masked_2022,
	title = {Masked {Autoencoders} for {Point} {Cloud} {Self}-supervised {Learning}},
	booktitle = {ECCV},
	author = {Pang, Yatian and Wang, Wenxiao and Tay, Francis E. H. and others},
	year = {2022},
}

@inproceedings{wei_3d_nodate,
	title = {{3D} {Test}-time {Adaptation} via {Graph} {Spectral} {Driven} {Point} {Shift}},
        booktitle = {ICCV},
	author = {Wei, Xin and Yang, Qin and Fang, Yijie and others},
	year = {2025},
}

@misc{kojima_robustifying_2022,
	title = {Robustifying {Vision} {Transformer} without {Retraining} from {Scratch} by {Test}-{Time} {Class}-{Conditional} {Feature} {Alignment}},
	booktitle = {IJCAI},
	author = {Kojima, Takeshi and Matsuo, Yutaka and Iwasawa, Yusuke},
	year = {2022},
}

@inproceedings{wang_tent_2021,
	title = {Tent: {Fully} {Test}-time {Adaptation} by {Entropy} {Minimization}},
	booktitle = {ICLR},
	author = {Wang, Dequan and Shelhamer, Evan and Liu, Shaoteng and Olshausen, Bruno and Darrell, Trevor},
	year = {2021},
}

@article{pan_survey_2010,
	title = {A {Survey} on {Transfer} {Learning}},
	journal = {IEEE TKDE},
	author = {Pan, Sinno Jialin and Yang, Qiang},
	year = {2010},
}

@InProceedings{finn_model-agnostic_2017,
	title = {Model-{Agnostic} {Meta}-{Learning} for {Fast} {Adaptation} of {Deep} {Networks}},
        booktitle = 	 {PMLR},
	author = {Finn, Chelsea and Abbeel, Pieter and Levine, Sergey},
	year = {2017},
}

@inproceedings{gidaris_unsupervised_2018,
	title = {Unsupervised {Representation} {Learning} by {Predicting} {Image} {Rotations}},
	booktitle = {ICLR},
	author = {Gidaris, Spyros and Singh, Praveer and Komodakis, Nikos},
	year = {2018},
}

@inproceedings{alet_tailoring_2021,
	title = {Tailoring: encoding inductive biases by optimizing unsupervised objectives at prediction time},
	shorttitle = {Tailoring},
        booktitle = {NeurIPS},
	author = {Alet, Ferran and Bauza, Maria and Kawaguchi, Kenji and others},
	year = {2021},
}

@inproceedings{sharafi_personalized_2026,
	title = {Personalized {Feature} {Translation} for {Expression} {Recognition}: {An} {Efficient} {Source}-{Free} {Domain} {Adaptation} {Method}},
	booktitle = {ICLR},
	author = {Sharafi, Masoumeh and Belharbi, Soufiane and Zeeshan, Muhammad Osama and others},
	year = {2026},
}

@InProceedings{yazdanpanah_revisiting_nodate,
    author    = {Yazdanpanah, Moslem and Bahri, Ali and Noori, Mehrdad and others},
    title     = {Revisiting Layer Normalization for Point Cloud Test Time Adaptation},
    booktitle = {WACV},
    year      = {2026},
}

@inproceedings{guichemerre_source-free_2024,
	title = {Source-{Free} {Domain} {Adaptation} of {Weakly}-{Supervised} {Object} {Localization} {Models} for {Histology}},
	booktitle = {CVPRw},
	author = {Guichemerre, Alexis and Belharbi, Soufiane and Mayet, Tsiry and others},
	year = {2024}
}

@article{zhang_deep_2026,
	title = {Deep {Learning}-{Based} {Point} {Cloud} {Registration}: {A} {Comprehensive} {Survey} and {Taxonomy}},
	journal = {IJCV},
	author = {Zhang, Yu-Xin and Gui, Jie and Yu, Baosheng and Cong, Xiaofeng and Gong, Xin and Tao, Wenbing and Tao, Dacheng},
	year = {2026},
}

@inproceedings{saltori2022gipso,
  title={GIPSO: Geometrically Informed Propagation for Online Adaptation in 3D LiDAR Segmentation},
  author={Saltori, Cristiano and Krivosheev, Evgeny and Lathuili{\'e}re, St{\'e}phane and others},
  booktitle={ECCV},
  year={2022},
}

@inproceedings{3600270.3602246,
author = {Gong, Taesik and Jeong, Jongheon and Kim, Taewon and others},
title = {NOTE: robust continual test-time adaptation against temporal correlation},
year = {2022},
booktitle = {NIPS},
}

@misc{hu_mixnorm_2021,
	title = {{MixNorm}: {Test}-{Time} {Adaptation} {Through} {Online} {Normalization} {Estimation}},
	publisher = {arXiv},
	author = {Hu, Xuefeng and Uzunbas, Gokhan and Chen, Sirius and others},
	year = {2021},
}

@article{welford1962,
  title   = {Note on a method for calculating corrected sums of squares and products},
  author  = {Welford, B P},
  journal = {Technometrics},
  year    = {1962},
}

@article{ma_visualization_2023,
	title = {Visualization, registration and tracking techniques for augmented reality guided surgery: a review},
	journal = {PMB},
	author = {Ma, Longfei and Huang, Tianqi and Wang, Jie and others},
	year = {2023},
}

@article{barrow_parametric_1977,
        title = {Parametric correspondence and chamfer matching : two new techniques for image matching},
	journal = {SRI internalional},
	author = {Barrow, H G and Tenenbaum J M and Bolles R C and others},
        year = {1977}
}

@inproceedings{gao_apcotta_2026,
	title = {{APCoTTA}: {Continual} {Test}-{Time} {Adaptation} for {Semantic} {Segmentation} of {Airborne} {LiDAR} {Point} {Clouds}},
	booktitle = {ISPRS},
	author = {Gao, Yuan and Xia, Shaobo and Nie, Sheng and others},
	year = {2026},
}

@inproceedings{ma_architecture-agnostic_2026,
	title = {Architecture-Agnostic Test-Time Adaptation via Backprop-Free Embedding Alignment},
	booktitle = {ICLR},
	author = {Ma, Xiao and Kwon, Young D and Zhou, Pan and Ma, Dong},
	year = {2026},
}

@inproceedings{wang_efficient_2021,
	title = {Efficient {Test} {Time} {Adapter} {Ensembling} for {Low}-resource {Language} {Varieties}},
	booktitle = {EMNLP},
	author = {Wang, Xinyi and Tsvetkov, Yulia and Ruder, Sebastian and others},
	year = {2021},
}

@inproceedings{zhang_domain-specific_2023,
    author = {Zhang, Yi-Fan and Wang, Jindong and Liang, Jian and others},
    title = {Domain-Specific Risk Minimization for Domain Generalization},
    year = {2023},
    booktitle = {KDD},
    }

\appendix
\section{Detailed TTA Methods for Registration}

\noindent \textbf{A) Point-TTA \cite{hatem_point-tta_2023}:}
The model parameters are $\theta = \{\theta^{sh}, \theta^{pri}, \theta^{aux}\}$, corresponding to a shared encoder, a primary registration head, and a self-supervised auxiliary head. During meta-training, an inner loop adapts the parameters using the auxiliary loss:
\begin{equation}
    \phi_b \leftarrow \theta - \gamma \cdot \nabla_{\theta} \mathcal{L}_{aux}(P^{b}_s, Q^b_s; \theta),
\end{equation}
and an outer loop updates $\theta$ so that the adapted parameters improve the primary task:
\begin{equation}
    \theta \leftarrow \theta - \beta \cdot \nabla_{\theta} \Bigl(\frac{1}{B} \sum_{b=1}^B \mathcal{L}_{pri}(P^{b}_s, Q^{b}_s, \mathcal{T}^{b}; \phi_{b}^{sh}, \phi_{b}^{pri})\Bigr),
\end{equation}
where the outer gradient is taken through the inner update, $\gamma$ and $\beta$ are the auxiliary and meta learning rates, and $B$ is the batch size. At test time, $K$ inner-loop updates are performed using the auxiliary loss. The shared encoder is adapted:
\begin{equation}
    \theta^{sh} \leftarrow \theta^{sh} - \gamma \cdot \nabla_{\theta^{sh}}\mathcal{L}_{aux}(P_t, Q_t; \theta).
\end{equation}
The auxiliary task analyzed here is point cloud reconstruction, using a Chamfer distance \cite{barrow_parametric_1977} loss. The encoder extracts features from $P$ and $Q$, and the decoder reconstructs the point clouds $P'$ and $Q'$:
\begin{equation}
    \mathcal{L}_{aux} = \tfrac{1}{2}\big(d_{CD}(P', P) + d_{CD}(Q', Q)\big),
\end{equation}
where $d_{CD}(A,B)$ is the Chamfer distance. Since the encoder is shared between the two clouds, this update is necessarily symmetric: unlike the other methods, Point-TTA cannot be restricted to the intraoperative cloud.

\noindent \textbf{B) Layer Normalization \cite{yazdanpanah_revisiting_nodate}:}
For layer normalization, the pre-affine ($h_{b,t}$) and post-affine ($y_{b,t}$) maps for batch $b$ and token $t$ are:
\begin{equation}
    h_{b,t} = \widehat{\mathrm{LN}}(F_{b,t}), \quad
    y_{b,t} = \Gamma_0^S h_{b,t} + \beta_0^S = \mathrm{LN}(F_{b,t}),
\end{equation}
with $\gamma_0^S, \beta_0^S$ the source post-affine parameters and $\Gamma_0^S = \mathrm{diag}(\gamma_0^S)$.

We use the source-moment variant of LN \cite{yazdanpanah_revisiting_nodate}, as the source-free variant relies on a [cls] token that PARE-Net \cite{yao_pare-net_2025} does not have. The method performs per-feature adaptation using the per-layer source post-affine moments $(\mu_{\star}^S, \Sigma_\star^S)$, with $\Sigma_\star^S = \mathrm{diag}(\sigma^{2,S}_{\star,1}, \dots, \sigma^{2,S}_{\star,d})$, and target pre-affine moments $(\hat{\mu}_h^T, (\hat{\sigma}_{h,j}^2)^T)$ estimated per sample (EP) or accumulated with the modified EMA of Eq. \ref{EMA} (CT). After computing the source statistics offline and the target statistics online, the output $y_{b,t}^{(T)}$ is adapted as:
\begin{align*}
    y_{\text{adapt}, b, t}^{(T)} = (y_{b,t}^{(T)} - \beta_0^S) \odot (\gamma^{\star} \oslash \gamma_0^S) + \beta^{\star}
\end{align*}
with $\gamma_j^{\star} = \mathrm{sgn}(\gamma_{0,j}^S)\sqrt{\frac{\sigma_{\star, j}^{2,S}}{(\hat{\sigma}_{h,j}^{2})^T+\delta}}$, $\Gamma^{\star} = \mathrm{diag}(\gamma^\star)$, $\beta^{\star} = \mu_{\star}^S - \Gamma^{\star}\hat{\mu}_h^T$, and $\delta > 0$ a small constant for numerical stability. In our modification, the source and target moments are stored and applied separately for the preoperative and intraoperative clouds.

\noindent \textbf{C1) Purge-Gate \cite{yazdanpanah_purge-gate_nodate}:}
For the source-prototype variant of \cite{yazdanpanah_purge-gate_nodate}, the divergence of the $j$-th target token embedding $F_j^T$ is the Mahalanobis distance to the source prototype $G = (\mu_S, \Sigma_S)$, with $\Sigma_S = \mathrm{diag}(\sigma_S^2)$:
\begin{equation}
    \Delta(F_j^T,G)
    = \sqrt{(F_j^T-\mu_S)^\intercal \Sigma_S^{-1}(F_j^T-\mu_S)}.
\end{equation}
We remove a fraction $L_{pg}$ of tokens of $Q$ with the largest $\Delta$ values. We use a ratio rather than the fixed number of tokens of \cite{yazdanpanah_purge-gate_nodate}, since the number of points varies across samples.

\noindent \textbf{C2) PEA \cite{ma_architecture-agnostic_2026}:}
PEA transforms the features $F$ using source and target mean and covariance statistics:
\begin{equation}
     F' = (F-\mu_t)\Sigma_t^{-1/2}\Sigma_s^{1/2} + \mu_s,
\end{equation}
where $\mu_s, \Sigma_s$ and $\mu_t, \Sigma_t$ are the source and target feature statistics. The covariance is decomposed as $\Sigma = V\Lambda V^{\top}$, with $V$ the eigenvectors and $\Lambda$ the eigenvalues, so that $\Sigma^{1/2} = V\Lambda^{1/2}V^{\top}$ and $\Sigma^{-1/2} = V\Lambda^{-1/2}V^{\top}$. We apply the alignment at a single level and use the aligned features directly, without the distance-aware weighting of \cite{ma_architecture-agnostic_2026}, which interpolates between the original and aligned features at every block. We also omit the entropy-based reset of the statistics and the test-time augmentation of \cite{ma_architecture-agnostic_2026}. For continual adaptation, the target statistics are updated online with an EMA of momentum $m$:
\begin{equation}
\label{EMA}
    \mu_t^{(i)} = (1-m)\mu_t^{(i-1)} + m\mu_b, \;
    \Sigma_t^{(i)} = (1-m)\Sigma_t^{(i-1)} + m\Sigma_b,
\end{equation}

where $\mu_B$ and $\Sigma_b$ are the statistics of the current batch. We replace $m$ with $\alpha_n = \max\!\left(\tfrac{n_b}{n},\, m\right)$, where $n_b$ is the number of target features in the current batch and $n$ the cumulative number observed. When $n_b/n > m$, $\alpha_n$ corresponds to the cumulative mean weight of the current batch, which is well defined even for a single-sample batch. Once $n_b/n$ falls below $m$, the update reverts to the constant momentum EMA of \cite{ma_architecture-agnostic_2026}.

\section{Detailed Results}
\subsection{Experimental methodology}
\noindent\textbf{Corruptions:} To characterize the geometric severity of each corruption, we measure the Density-aware Chamfer Distance (DCD) \cite{wu_density-aware_2021} between the intraoperative cloud and the ground-truth-aligned preoperative cloud. Table \ref{tab:dcd_corruptions} shows that, relative to clean data, noise corruptions induce a much larger discrepancy on P2ILReg-Syn than on P2P-Syn, whereas density and occlusion corruptions have a limited effect on P2ILReg-Syn. This is consistent with Table \ref{tab:results_rmse}, where P2ILReg-Syn errors are dominated by noise corruptions.

\begin{table}[H]
\centering
\setlength{\tabcolsep}{3pt}
\begin{tabular}{c|cccccccc}
\toprule
clean & uni & gauss & backg & impul
& g-d-dec & l-d-dec & cut & oclsion \\
\midrule
\multicolumn{9}{c}{\textbf{P2P-Syn}}\\
\midrule
0.769 & 0.807 & 0.811 & 0.782 & 0.786 & 0.788 & 0.793 & 0.785 & 0.789\\
\midrule
\multicolumn{9}{c}{\textbf{P2ILReg-Syn}}\\
\midrule
0.805 & 0.937 & 0.949 & 0.844 & 0.859 & 0.810 & 0.809 & 0.810 & 0.809\\
\bottomrule
\end{tabular}%
\caption{DCD between the intraoperative and ground-truth-aligned preoperative point clouds on clean and corrupted data, averaged over five severity levels.}
\label{tab:dcd_corruptions}
\end{table}

\noindent \textbf{Additional implementation details:}
For PARE-Net \cite{yao_pare-net_2025}, the checkpoint with the lowest validation RMSE is selected for registration. Source training took 2 hours and 7 minutes on P2P-Syn \cite{yang_resolving_2025}, and 2 days, 8 hours and 5 minutes on P2ILReg \cite{zhou_landmark-free_2025}. For Point-TTA \cite{hatem_point-tta_2023} on P2P-Syn, the joint training took 5 hours and 1 minute, and the meta-training 7 hours and 43 minutes.

\subsection{Results}

\noindent \textbf{Multi-level alignment:}
In our implementation, PEA \cite{ma_architecture-agnostic_2026} aligns the intraoperative features at a single level. The original method aligns the embeddings progressively at every block. We therefore evaluate whether a multi-level alignment (PEA-ML) further improves registration. As in the single-level version, the alignment is restricted to the intraoperative features, leaving the preoperative embeddings unchanged.

\begin{figure}
    \centering
    \includegraphics[width=1.0\linewidth]{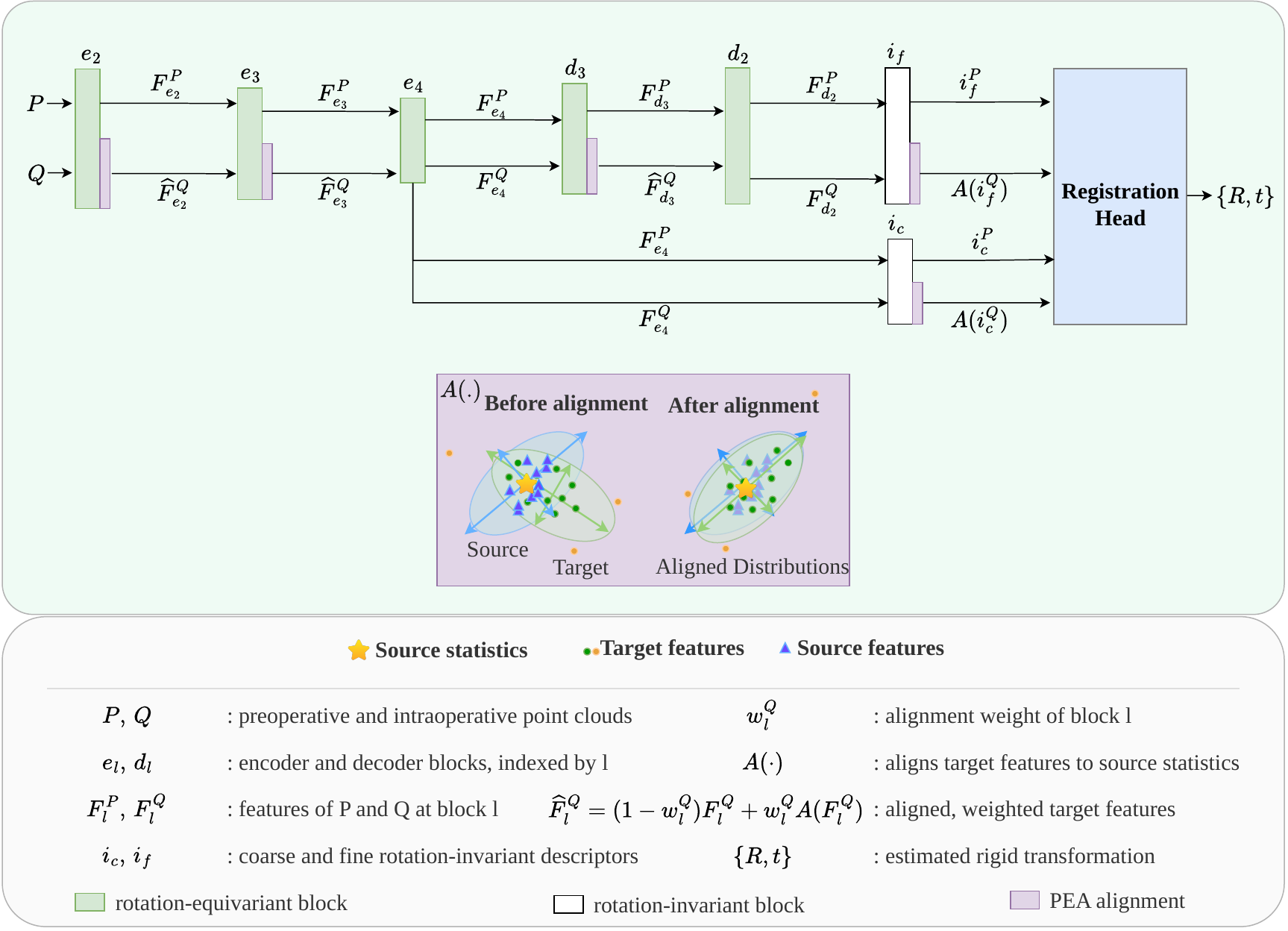}
    \caption{Multi-level PEA (PEA-ML). Target intraoperative features are progressively aligned with source statistics at multiple levels.}
    \label{fig:PEA_bckb}
\end{figure}

In PEA \cite{ma_architecture-agnostic_2026}, a first forward pass estimates the domain shift at different blocks. For the rotation-equivariant blocks, these shifts are min-max normalized to obtain alignment weights $w^Q_l \in [0,1]$, which control the contribution of the aligned features. The shift of the rotation-invariant blocks $i_c$ and $i_f$ are larger, and including them in the normalization would compress the weights of the preceding blocks toward zero. We therefore exclude $i_c$ and $i_f$ from the normalization and set their alignment to 1. Fig. \ref{fig:PEA_bckb} illustrates PEA-ML.

\begin{table}[]
    \centering
    \small
    \begin{tabular}{|c|c|c|c|c|}
        \hline
        Method & P2P-Syn & P2ILReg-Syn & \multicolumn{2}{c|}{P2ILReg-Real} \\
        & RMSE (mm)\,$\downarrow$ & RMSE (mm)\,$\downarrow$ & CD (mm)\,$\downarrow$& Dice $\uparrow$ \\
        \hline
        Source only & 15.47 & 38.70 & 4.26 & 71.87 \\
        \hline 
        PEA (EP) & 13.83 & 36.90 & 4.27 & 72.85 \\
        PEA (CT) & 13.98 & 36.77 & 4.20 & 72.18 \\
        \hline 
        PEA-ML (EP) & \textbf{11.16} & \textbf{31.88} & \textbf{3.87} & \textbf{74.46} \\
        \hline
    \end{tabular}
    \caption{Registration performance with single-level PEA and PEA-ML.}
    \label{tab:MA}
\end{table}

Table \ref{tab:MA} shows that multi-level alignment improves registration on all three datasets. On P2P-Syn and P2ILReg-Syn, RMSE decreases from 13.83 to 11.16 mm and from 36.90 to 31.88 mm, respectively, compared with episodic single-level PEA. On P2ILReg-Real, CD decreases from 4.27 to 3.87 mm and Dice increases from 72.85 to 74.46. The average inference time increases slightly, from 189 ms for single-level alignment to 196 ms for multi-level alignment, due to the additional forward pass for weight estimation and the covariance decompositions performed at multiple levels.

\noindent \textbf{Inlier Ratio (IR):}
For Purge-Gate, we replace the entropy criterion with IR. To assess whether IR provides a suitable unsupervised criterion for registration, we measure its Spearman rank correlation with CD on all datasets, RMSE on synthetic data and Dice on real data: 
\begin{equation}
\rho \;=\; \frac{\operatorname{cov}\!\left(R(s),\, R(e)\right)}
{\sigma_{R(s)}\,\sigma_{R(e)}},
\label{eq:spearman}
\end{equation}
where $R(\cdot)$ denotes the rank, $s_i$ the predicted inlier ratio $\mathrm{IR}(\hat{T},\hat{C})$ and $e_i$ the corresponding evaluation metric for a pair $i$. Correlations are computed over all valid pairs, excluding cases for which no transformation hypothesis is produced. A negative correlation is expected for CD and RMSE, and a positive correlation is expected for Dice. 

In order to evaluate the variation of IR across samples, we also report its 5th and 95th percentiles ($P_{5}$ and $P_{95}$) and the interpercentile range $\text{IPR} = P_{95} - P_{5}$.

Results are reported in Table \ref{tab:spearman}.

\begin{table}[]
    \centering
    \begin{tabular}{|c|c|c|c|c|c|}
        \hline
        Dataset & IPR & IR[ $P_{5}$, $P_{95}$] & $\rho_{\text{CD}}$ &  $\rho_{\text{RMSE}}$ &  $\rho_{\text{Dice}}$ \\
        \hline
        P2P-Syn &  0.810 & [0.118, 0.928] & $-$0.477 & $-$0.831 & - \\
        P2ILReg-Syn & 0.636 &[0.011, 0.647] & $-$0.402 & $-$0.879 & - \\
        P2ILReg-Real & 0.081 &  [0.022, 0.103] & $-$0.255 & - & 0.303 \\
        \hline
    \end{tabular}
    \caption{Spearman rank correlation between IR and the registration metrics. The IR 5th–95th percentile range and interpercentile range (IPR) are also reported.}
    \label{tab:spearman}
\end{table}

Table \ref{tab:spearman} shows a strong negative correlation between IR and RMSE on P2P-Syn and P2ILReg-Syn (-0.831 and -0.879). The correlation with CD is weaker on the three datasets, particularly on P2ILReg-Real. On real data, IR is also weakly positively correlated with Dice. The weaker correlations on P2ILReg-Real may partly result from the smaller variation of IR, with an IPR of 0.081 compared with 0.810 and 0.636 on P2P-Syn and P2ILReg-Syn, respectively.

\begin{figure}
    \centering
    \includegraphics[width=1.0\linewidth]{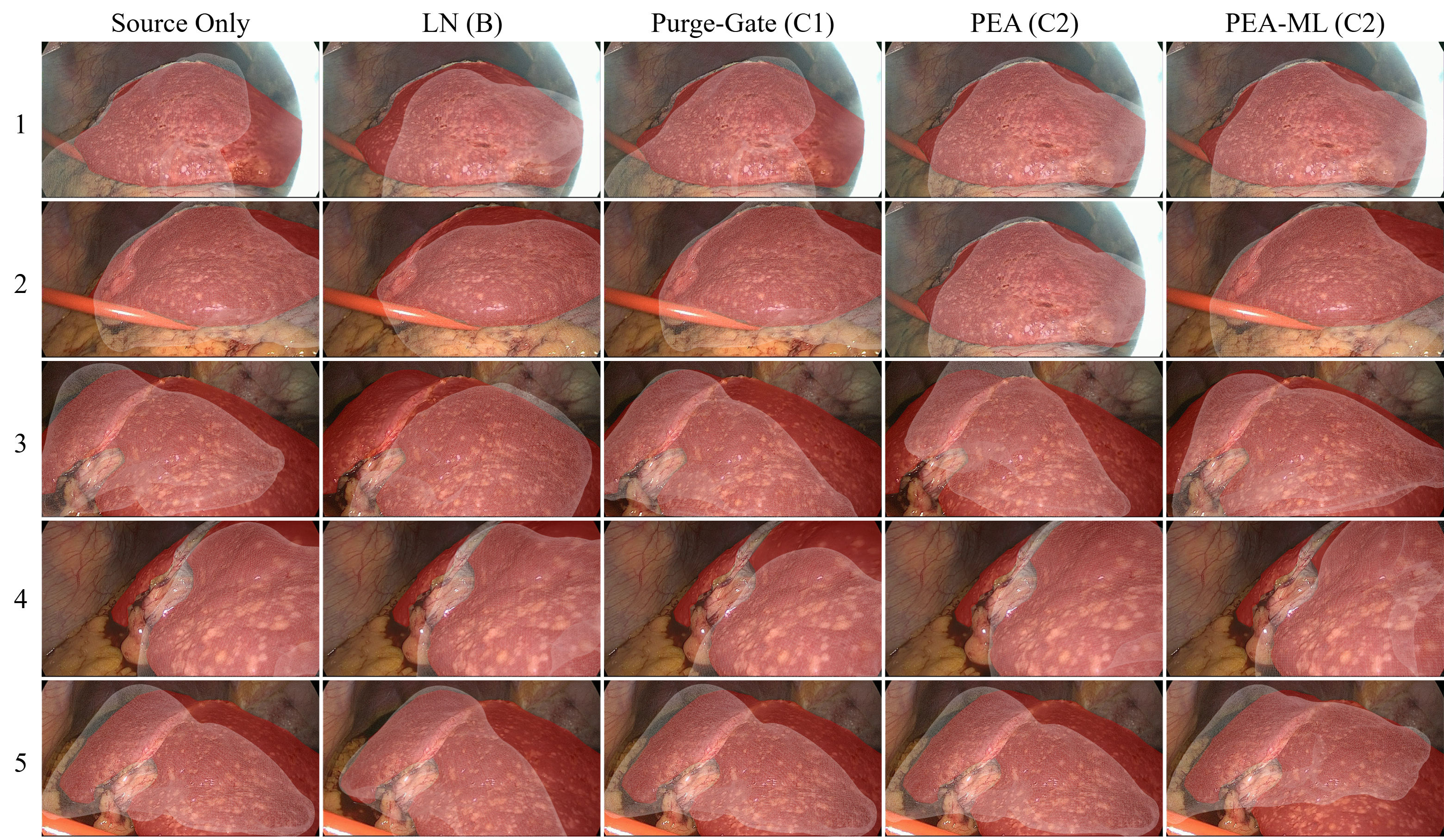}
    \caption{Registration results on five non-consecutive frames from one patient of P2ILReg-Real. Episodic adaptation is shown for LN, PEA, and PEA-ML. The white mesh represents the registered preoperative organ, and the red region indicates the intraoperative organ segmentation.}
    \label{fig:vizualisations}
\end{figure}

\begin{figure}
    \centering
    \includegraphics[width=1.0\linewidth]{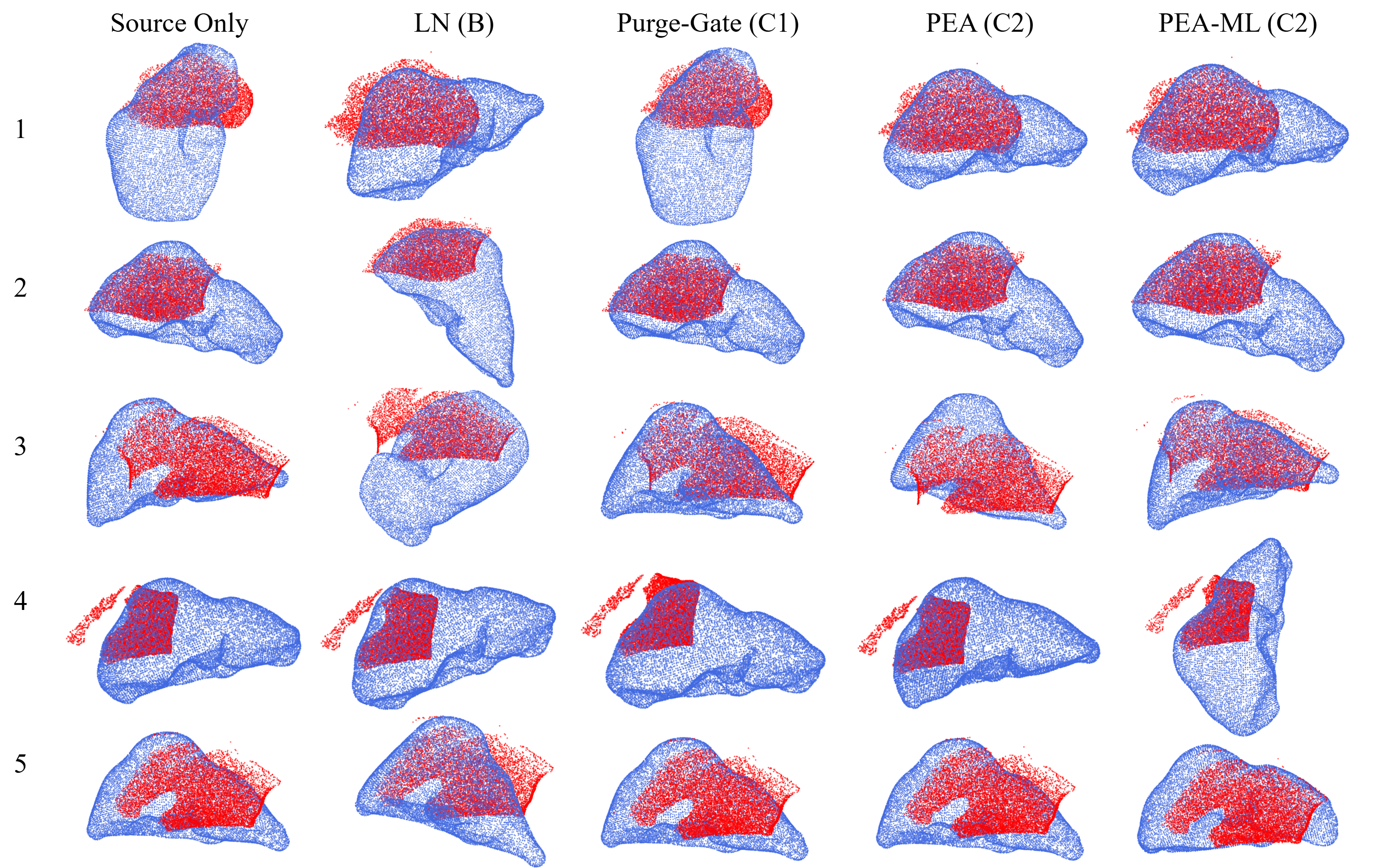}
    \caption{Point-cloud registration results for the same five frames. The intraoperative and transformed preoperative point clouds are shown in red and blue, respectively.}
    \label{fig:vizualisations_matches}
\end{figure}

\noindent \textbf{Visual results:}
We visualize the registrations obtained on five non-consecutive frames from one patient of the P2ILReg-Real test set, shown in Figs. \ref{fig:vizualisations} and \ref{fig:vizualisations_matches}. Since ground-truth transformations are unavailable, differences between most methods remain difficult to assess visually. LN produces the clearest degradation, with the registered mesh visibly deviating from the organ segmentation.

\newpage

\end{document}